\documentclass[5p,nopreprintline]{elsarticle}
\usepackage{amsmath,amssymb,amsfonts}
\usepackage[ruled,linesnumbered]{algorithm2e}
\SetKwInput{KwInput}{Input}                
\SetKwInput{KwOutput}{Output}              
\usepackage{graphicx}
\usepackage{textcomp}
\usepackage{physics}
\usepackage{siunitx}
\usepackage{xcolor}
\usepackage{amsmath}
\usepackage{booktabs}
\usepackage{multirow}
\def\BibTeX{{\rm B\kern-.05em{\sc i\kern-.025em b}\kern-.08em
    T\kern-.1667em\lower.7ex\hbox{E}\kern-.125emX}}

\begin{document}

\begin{frontmatter}

\title{Machine Learning Driven Global Optimisation Framework for Analog Circuit Design}

\author[inst1]{Ria Rashid\corref{cor1}}
\cortext[cor1]{Corresponding Author}
\ead{ria183422005@iitgoa.ac.in}
\author[inst1]{Komala Krishna}
\ead{krishna183422003@iitgoa.ac.in}
\author[inst2]{Clint Pazhayidam George}
\ead{clint@iitgoa.ac.in}
\author[inst1]{Nandakumar Nambath}
\ead{npnandakumar@iitgoa.ac.in}

\affiliation[inst1]{organization={School of Electrical Sciences, Indian Institute of Technology Goa},
            city={Ponda},
            postcode={403401}, 
            state={Goa},
            country={India}}

\affiliation[inst2]{organization={School of Mathematics and Computer Science, Indian Institute of Technology Goa},
            city={Ponda},
            postcode={403401}, 
            state={Goa},
            country={India}}

\begin{abstract}

  We propose a machine learning-driven optimisation framework for analog circuit design in this paper. Machine learning based global offline surrogate models, with the circuit design parameters as the input, are built in the design space for the analog circuits under study and is used to guide the optimisation algorithm towards an optimal circuit design, resulting in faster convergence and reduced number of spice simulations. Multi-layer perceptron and random forest regressors are employed to predict the required design specifications of the analog circuit. Multi-layer perceptron classifiers are used to predict the saturation condition of each transistor in the circuit. We validate the proposed framework using three circuit topologies--a bandgap reference, a folded cascode operational amplifier, and a two-stage operational amplifier. The simulation results show better optimum values and lower standard deviations for fitness functions after convergence, with a reduction in spice calls by 56\%, 59\%, and 83\% when compared with standard approaches in the three test cases considered in the study.
 
\end{abstract}


\end{frontmatter}
\section{Introduction}

  The present-day electronic industry uses more and more integrated analog and digital blocks on monolithic mixed-signal system-on-a-chip (SoC). The main bottleneck to the rapid development cycles of SoCs is the lack of automation in the analog design process. This is primarily because of the complexities present in analog circuit design \cite{1}. Digital circuit design, on the other hand, is heavily automated. This necessitates the development of new, robust analog design automation tools \cite{2,3}. 

  Multiple studies have been reported for the automated sizing of analog circuits \cite{4, 48, 105, 106}, which are typically classified into equation-based and simulation-based \cite{5,6}. In equation-based approaches \cite{6,7}, analytical expressions are used to model the different circuit specifications with respect to design parameters. Since the models of the state-of-the-art transistors are highly complex, developing an accurate circuit model becomes quite challenging. Various higher-order effects are thus ignored while framing the required equations for circuit performance evaluation. As a result, the optimal design achieved by these methods is often found to be inadequate. 
  
  In simulation-based approaches \cite{8,51,107,108,109,113}, optimisation algorithms find the design parameters of the considered analog circuit to meet the required specifications with the help of any electronic design automation (EDA) tool, including spice \cite{31}. Combined with the ability of spice simulations to predict the performance metrics of any analog circuits, this method can be applied to any complex analog circuit without needing to develop accurate mathematical models of the circuit under study. Different methods have been reported in the literature with various 
  simulation-based techniques for analog circuit optimisation. Self-adaptive multiple starting point optimisations \cite{9}, simulated annealing \cite{10, 49}, Bayesian optimisation \cite{11,111}, artificial intelligence-based approach \cite{12}, shrinking circles technique \cite{13}, machine learning-based optimisation methods \cite{14,112}, and evolutionary algorithms \cite{15,16,17, 50} are a few methods reported for analog circuit design optimisation. A variety of evolutionary global optimisation algorithms such as differential evolution (DE) \cite{18}, genetic algorithm (GA) \cite{19,110}, artificial bee colony (ABC) algorithm \cite{34}, butterfly algorithm \cite{103} and particle swarm optimisation (PSO) \cite{20} have been reported for analog circuit design because of their robustness. In our prior work \cite{34}, a performance evaluation of modified versions of ABC, GA, grey wolf optimisation (GWO) and PSO was carried out for the optimisation of two operational amplifier (op-amp) topologies. The modified versions of these algorithms showed faster convergence to better optimal values with a reduced number of spice simulations when compared to standard versions.

  GA has found extensive applications in studies where parallel computation is utilised \cite{21, 22}. Since every individual in the GA population can be evaluated independently from the others, the implementation of parallel computation becomes straightforward. For an efficient global exploration of the design space, GA usually requires a large population size, leading to more spice simulations. This makes GA more computationally expensive, which can be a significant constraint when complex circuits with stringent specifications are considered \cite{23}.

  One way to overcome the computation-intensive circuit simulation is by creating surrogate models for different circuit parameters. These surrogate models are approximate models of circuit simulations that can be built using different techniques and can be used to replace circuit simulations during optimisation. One main advantage of such a model is that it can be saved and reused for different optimisation runs. Machine learning (ML)-based surrogate models have been reported in the literature for analog circuit optimisation \cite{24, 39, 40}. 
 
  ML-based regression models have been used in different studies to predict the various circuit specifications for analog circuits and are used instead of the spice simulations to predict the feasibility of a design attained by the optimisation algorithm. Different methods such as artificial neural network (ANN) \cite{40}, posynomial models \cite{40,41}, support vector machines \cite{44}, random forest (RF) \cite{42}, k-nearest neighbours \cite{43}, and deep neural network \cite{55} have been employed in analog design optimisation. A circuit-connectivity-inspired ANN was proposed in \cite{46}, reducing the required data set volume for a specified target accuracy in analog circuit design. In \cite{24}, an ANN-based methodology for generating fast and efficient models for estimating the performance parameters of complementary metal-oxide semiconductor (CMOS) op-amps was presented, where simulation results demonstrated the efficiency of the proposed method in the performance estimation of analog circuits. A GA-based global optimisation engine and an ANN-based local optimisation engine for analog circuit optimisation were presented in \cite{25}. This study implemented parallel computation to train the ANN models during the final local optimisation search, and this proved to be faster with comparable results when compared with the local optimisation approach using spice calls. In \cite{26}, a local surrogate-based parallel optimisation has been proposed for analog design. This method showed better optimisation ability than the parallel DE algorithm and the state-of-the-art surrogate-based optimisation methods. An ML-based global optimisation approach with a new candidate design ranking method and an ANN model construction method for the analog circuits was reported in \cite{27} and validated using two amplifiers and a comparator with complete design specifications. An efficient surrogate-assisted constrained multi-objective evolutionary algorithm for analog circuit sizing via self-adaptive incremental learning was presented in \cite{28}. Simulation results on three real-world circuits showed the method's superiority in reducing the total optimisation time compared to Bayesian optimisation. 

  \begin{figure*}[!t]
    \centering
    \includegraphics[width=0.9\textwidth]{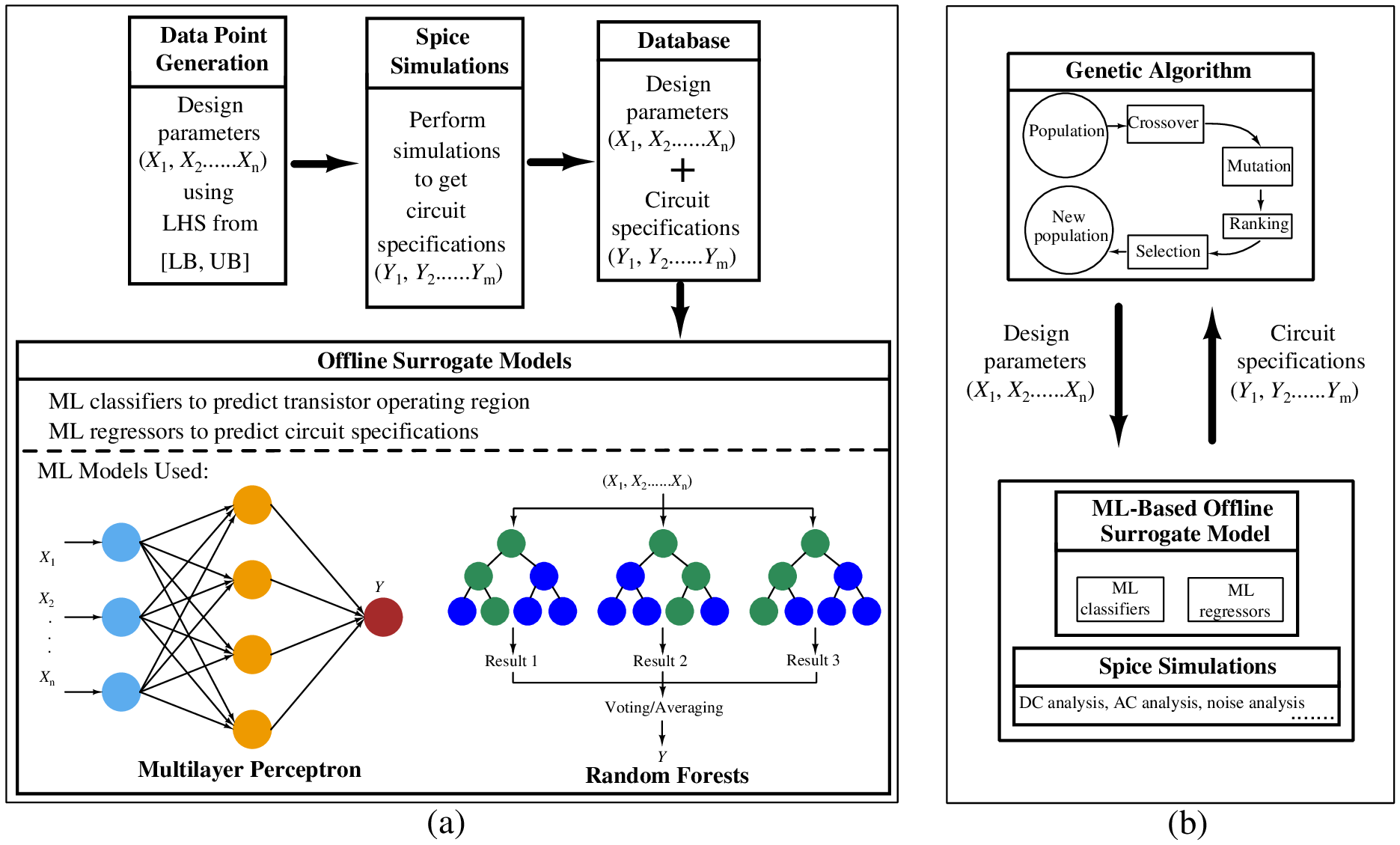}
    \caption{Illustration of the proposed global optimisation framework powered by machine learning algorithms. Block (a) shows the database generation and the development of offline ML models. Block (b) shows the interaction of optimisation algorithms with the developed ML models and spice simulations.}
    \label{fig4}
  \end{figure*}

  Surrogate models with high accuracy usually need a large number of circuit simulations. If the model is not accurate enough, it will affect the optimisation efficiency \cite{29}. Surrogate-based optimisation (SBO) can be divided into offline SBO \cite{30} and online SBO \cite{31}. In offline SBO, an offline surrogate model is built initially with an extensive database. This model is used to replace the circuit simulations. In online SBO, the model is built on a small database initially but continuously retrained as and when new data points are added to the database during the optimisation run. Since the inaccuracy of a surrogate model is inevitable in both these cases, the model may erroneously lead the optimisation engine to a local or a false optimum. In offline SBO, during the local optimisation search, the constructed surrogate model could lead the optimisation engine to a false optimum as there is no additional validation with spice simulations. In the case of online SBO, the child population in each generation is ranked first by the surrogate model. Only the best candidate is subjected to spice simulations and added to the database. This will be an issue when none of the individuals in the child population satisfies all the constraints. Since the optimisation engine starts with a limited database, this is a major concern in the initial iterations, especially when a complex circuit with stringent specifications is considered. A recent study \cite{32} has reported that no matter how accurate a machine learning model is for a single specific problem, a simple combination of evolutionary and machine learning methods cannot achieve the desired results without regular supervision by any physics-based tool. The inaccuracy of GA with a neural network model in converging to a global optimum without regular supervision using a physics-based tool has been proven with experimental results in \cite{32}. A framework with an ML model and regular supervision with a physics-based tool is reported to have the highest optimisation efficiency. It would be productive to see how such a framework applies to analog circuit optimisation problems.

  In this study, we use an ML surrogate model-based optimisation technique with regular supervision using a circuit simulation tool for analog circuit design. While designing an analog circuit, it is essential that the right operating region of the transistors is ensured for the circuit to work properly. For example, in a two-stage Miller compensated op-amp, which is one of the test cases considered in this study, it is vital to ensure that all the transistors are in the saturation region. We propose ML-based classification models in this work to predict whether the transistors are in the right region of operation. All the other relevant circuit specifications, such as gain, phase margin, bandwidth, etc., are modelled using ML-based regression models. We propose a spice simulation-guided ML surrogate-based optimisation framework in which ML surrogate models are used to check the feasibility of a new design point before invoking spice calls. We use GA as the optimisation algorithm because of its robustness in parallel implementation \cite{25}. 

  We analysed the performance of the proposed framework using three different analog circuits, namely, a bandgap reference (BGR) circuit, a folded cascode op-amp (FCOA), and a two-stage Miller compensated op-amp (TSMCOA). The effect of the novel techniques, namely, modified GA with ML-aided saturation prediction (MGA-MLSP) and modified GA with ML-aided saturation and constraint prediction (MGA-MLSCP) when compared with modified GA (MGA) \cite{34} and standard GA (SGA) in reducing the spice simulations and precision improvement is summarised in Table~\ref{table22}. In the spice calls comparison, the column MGA-MLSCP shows the mean value of the number of spice simulations invoked during optimisation by MGA-MLSCP in multiple runs. The other columns reflect the spice simulations invoked by other methods in relation to MGA-MLSCP. MGA-MLSP and MGA-MLSCP take considerably fewer spice simulations than the other two methods. The ML-assisted algorithms perform the best in the test case of TSMCOA, where the design specifications are more stringent. We consider the input common-mode range (ICMR) specification in the TSMCOA test case, and the optimised design has to satisfy all the other circuit specifications in this range. This shows that MGA-MLSP and MGA-MLSCP perform better in cases where specifications are more stringent. When comparing the standard deviation (S.D.) of the optimal value obtained for multiple runs by the four approaches, MGA-MLSP and MGA-MLSCP have the lowest values for all three test cases, showing an improved precision. It can be inferred from the results in Table~\ref{table22} that the proposed ML-assisted algorithms perform significantly better in terms of computational time and precision than SGA and MGA (More details on simulations in Section \ref{section3}). 

    \begin{table}[!t]
    \begin{center}
    \caption{Comparison of the number of spice simulations and improvement in precision by using the proposed framework with those of the competing frameworks on the three test circuits (Details in Section \ref{section3}).}
    \label{table22}
    \scalebox{1}{
      \begin{tabular}[!t]{ccccc}
      \toprule     
    \multicolumn{5}{c}{\textbf{Comparison in terms spice calls}}\\
    \midrule
    {\textbf{Test Case}}&\textbf{MGA-}&\textbf{MGA-}&{\textbf{MGA}}&{\textbf{SGA}}\\
    {\textbf{}}&\textbf{MLSCP}&\textbf{MLSP}&{\textbf{}}&{\textbf{}}\\
      \midrule
        BGR & 8178& 1.1$\times$ &1.4$\times$ & 2.3$\times$\\
        FCOA &8639 &1.4$\times$ & 2.4$\times$ &2.5$\times$ \\
        TSMCOA&6994&1.5$\times$ & 3.6$\times$ & 6.2$\times$\\
      \midrule
      \multicolumn{5}{c}{\textbf{Comparison in terms of error (S.D.)}}\\
        \midrule
      \textbf{Test Case}&\textbf{MGA-}&\textbf{MGA-}&\textbf{MGA}&{\textbf{SGA}}\\
      \textbf{}&\textbf{MLSCP}&\textbf{MLSP}&\textbf{}&{\textbf{}}\\
      \midrule
        BGR & 0.294 &0.456& 0.554 & 2.421\\
        FCOA &0.022&0.041&0.069&0.660\\
        TSMCOA&0.001&0.002&0.005&0.022\\
        \bottomrule
      \end{tabular}} 
    \end{center}
  \end{table}

  The main contributions of the study are summarized below. (i) We demonstrate the application of machine learning-based offline surrogate models to achieve an optimal design for analog circuits under constant EDA tool supervision. (ii) We also show that machine learning classifiers efficiently predict the operating region of transistors in the analog circuits under study, with very high accuracy. It has significantly reduced the expensive circuit simulations undertaken during optimisation runs in test circuits considered in this study.

  The rest of this paper is organised as follows. Section \ref{section2} details the ML-driven analog circuit design framework proposed in this study. The formulation of the optimisation problem for the three test circuits and the simulation results are discussed in Section \ref{section3}. The concluding remarks are presented in Section \ref{section4}. 

\section{Machine Learning-Driven Analog Circuit Design Framework}
\label{section2}

  {
 \begin{algorithm}[]
\DontPrintSemicolon
 \BlankLine
  \KwInput{$N$, $D$, $gen_{\max}$, $LB$, $UB$ }
  \KwOutput{Best individual (Optimal solution)}
  \BlankLine
   \For{$i\gets1$ \KwTo $N$}{\While{true}{
    Form an individual, $\vb*x_i$, by picking values randomly from [$LB$, $UB$]\;
    \If{Feasibility check($\vb*x_i$) = Passed}{Break.}
    }
    }
  \For{$gen\gets1$ \KwTo $gen_{\max}$}{
    \For{$i\gets1$ \KwTo $N$}{
    \While{true}{
    Randomly pick two parents and the crossover point.\;
    Generate crossover offspring, $\vb*x_{c,i}$.\;
    \If{Feasibility check($\vb*x_{c,i}$) = Passed }{Break.}
    }
    \While{true}{
    Find search space parameter, $\alpha$ and outer bounds for mutation.\;
    Randomly pick $mutation_{count}$ and $mutation_{no}$.\;
    Generate mutated crossed offspring, $\vb*x_{cm,i}$.\;
    \If{Feasibility check($\vb*x_{cm,i}$) = Passed }{Break.}
    }
    \While{true}{
    Find search space parameter, $\alpha$ and outer bounds for mutation.\;
    Randomly pick a parent, $mutation_{count}$ and $mutation_{no}$.\;
    Generate mutated parent offspring, $\vb*x_{pm,i}$.\;
    \If{Feasibility check($\vb*x_{pm,i}$) = Passed }{Break.}
    }
    }
    Pool together the parents, $\vb*X$ and offspring, $\vb*X_{c}$, $\vb*X_{cm}$, and $\vb*X_{pm}$, to form $\vb*X_{new}$.\;
    Sort $\vb*X_{new}$ according to fitness function value.\;
    Select fittest $N$ individuals to form new population, $\vb*X$.\;
    Update best individual.\;
    }
\textbf{End Function}
\caption{Modified GA (MGA) with saturation and constraint prediction (MGA-MLSCP).}
\label{alg1}
\end{algorithm}}%

\begin{algorithm}[!t]
\SetAlgoLined
\DontPrintSemicolon
\KwIn{ $\vb*x$ (new individual generated by GA)}    
\KwOut{Passed or Not Passed}
    \SetKwFunction{FMain}{Feasibility check}
    \SetKwProg{Fn}{Function}{:}{}
    \Fn{\FMain{$\vb*x$}}{
        \eIf{ML classifier saturation check and regressor specification check passes}{
        Perform spice simulation\\
        \eIf{Spice check passes}{Passed}{Not passed}
        }{Not passed}
}
\textbf{End Function}
\caption{Feasibility check}
\label{alg2}
\end{algorithm}

  Analog circuit design involves a trade-off between various circuit parameters such as gain, power, speed, and supply voltage. As such, analog design can be considered a multidimensional constrained optimisation problem. To frame the analog circuit optimisation problem, the circuit specification that needs to be optimised is regarded as the objective function, and the other specifications are modeled as the constraints for the problem. The design parameters of the analog circuit are considered as the decision variables, and their bounds form the search space for the optimisation problem. Any suitable optimisation algorithm can then be applied in the search space to find the optimum solution which satisfies all the constraints. In this study, we present an ML-aided framework for analog circuit design (Illustration in Fig.~\ref{fig4}). ML models predict circuit specifications and transistor saturation conditions for every considered design during the optimisation run.

  For ML-based optimisation of an analog circuit, a database needs to be developed for training the ML-based surrogate models (Fig.~\ref{fig4}(a)). The database must contain information of all relevant design parameter values of the analog circuit and the resulting circuit specification values, including the operating region of all the transistors in the circuit. The circuit specification, which is to be regarded as the objective function to be optimised, and the circuit specifications to be considered as constraints are selected. Latin hypercube sampling (LHS) from the pyDOE package in Python is used to generate the initial database within the prescribed limits of the search space, with parallelised \texttt{ngspice} simulations. The database is populated with saturation information of each of the transistors in the circuit, as well as the constraint circuit specifications for all the design points generated using LHS. We train ML-based offline surrogate models to predict various parameters of an analog circuit using this database. We build ML regression models to predict the relevant circuit specifications and ML classification models to predict the transistors' saturation conditions for the analog circuit.

  Depending on the circuit, ML regression models are built separately for each circuit specification, such as gain, phase margin, bandwidth, etc.. These models are tuned independently for each circuit specification for the best performance. The transistors must meet the saturation condition for the proper functioning of analog circuits. During an optimisation run with GA and parallel spice simulations, without using any ML model predictions, we observe that the saturation condition fails many times, leading to a substantial increase in total spice simulations. We thus use ML classification models to predict whether each transistor is in saturation, which can significantly reduce the spice simulations invoked during the optimisation run. ML classification models have been used for the saturation prediction of the transistors in all the test cases. The number of spice simulations required for building the database is less when compared with the number of spice simulations saved in the optimisation process. After using different ML models, we find multi-layer perceptron (MLP) and random forest (RF) regression models most suitable to predict the circuit specifications, and we use MLP classification models to predict the saturation condition of transistors. 

  The optimisation algorithm interacts with the learned ML-based offline surrogate models along with spice simulations to give an optimal design for the analog circuit under study (Fig.~\ref{fig4}(b)). We use an adapted and parallelized version of GA, MGA \cite{34} as the optimisation algorithm with \texttt{ngspice} as the circuit simulator. Spice simulations, along with ML model predictions, guide the GA to the global optimal solution in the search space. The aim of using such a framework for analog circuit optimisation is to increase the optimisation efficiency and to reduce the spice simulations undertaken during the optimisation run. 
  
  The results using two variants of the proposed framework, namely, MGA-MLSP and MGA-MLSCP, are presented in this work and compared with respect to the optimal solution obtained and the number of spice simulations needed during optimisation. In MGA-MLSP, whenever a new individual is created in the current population in GA, the saturation condition of each transistor in the circuit is predicted using ML classifiers. The spice simulations are invoked only if all transistors are predicted to be in saturation. Similarly, in MGA-MLSCP, whenever a new individual is formed in GA, along with the saturation condition by ML classifiers, the constraints are also predicted by ML regression models. Spice simulations are invoked if the ML models predict that both the saturation and constraint conditions are being met by the particular design. Algorithms \ref{alg1} and \ref{alg2} give details of the feasibility check undertaken using ML models and MGA-MLSCP algorithms. Note that $N$, $D$, $gen_{\max}$, $LB$, and $UB$ represent the population size, the dimension of the search space, the maximum number of generations, lower bound and upper bound of the decision variables, respectively. We describe the optimisation algorithm and the ML models used in this study next.

\subsection{Genetic Algorithm}

  GA is a metaheuristic optimisation method based on the theory of natural evolution, introduced by John Holland in 1971 \cite{33}. The concept of survival of the fittest mainly governs this algorithm. When applied to an optimisation problem, the algorithm imitates the process of natural selection to find the fittest individuals in a population. These fittest individuals are then selected for reproduction to create new offspring for the next generation. 
  
  The GA implemented in this study is a modified version adapted from \cite{34}. The modified GA has a search space parameter $\alpha$, which decreases linearly over the generations, thereby limiting the search space of the algorithm and aiding in faster convergence. After selection based on the fitness scores of all the individuals, parents are randomly selected to create offspring in each generation using a crossover operation, with the crossover point selected randomly. For each crossover offspring created, a mutated offspring is also generated. The number of genes to be mutated and which genes to be mutated are picked randomly for the mutation operation. Some parents are also randomly picked to create mutated offspring. The parents, crossover offspring, crossover mutated offspring, and parent mutated offspring are then pooled together, and the fittest among them are carried over to the next generation. The process is repeated till the stopping criteria is met. A detailed explanation and pseudo code of MGA is given in \cite{34}.

\subsection{Machine Learning Models}

We now discuss two machine learning models, multi-layer Perceptrons (MLP) and random forests (RF), employed in this study. We use MLP and RF regression models to estimate circuit specifications such as phase margin and gain. MLP classifiers are used to predict transistor saturation conditions. We also studied other standard machine learning classifiers, such as $k$-nearest neighbors and support vector machines. We discard them in the discussion as we found that MLP and RF worked more efficiently in our empirical study.
  
\paragraph{Multi-Layer Perceptrons}

  MLP is a supervised learning algorithm that is a fully connected feed-forward ANN and is used in the literature for analog circuit optimisation \cite{53, 54}. We use the training data to train the model or the function, $f(.): R^m \xrightarrow{} R^n$, where $m$ and $n$ are the input and output dimensions, respectively. For a given set of features and a target, MLP can model a non-linear function for both classification and regression. It has an input layer, an output layer, and one or more middle layers, called hidden layers. Every input feature has a neuron or node in the input layer. Similarly, there is a node for every output in the output layer. Each hidden layer can have any number of nodes. The nodes in the input layer take the input and forward it to every node in the first hidden layer. The hidden layers transform the input using a weighted linear summation and a non-linear activation function. This information is then passed onto the nodes in the output layer. In this study, both MLP regression models and classification models have been used for prediction. In this study, we have used MLP implementations from \texttt{scikit-learn} \cite{35}. The MLP implementation trains using backpropagation, and more details can be found in \cite{35}.

\paragraph{Random Forests}
 
  RF is a popular supervised machine learning model using an ensemble learning technique, and has been reported for the design of analog circuits \cite{43}. It can be modeled for classification as well as regression problems. RF works by constructing several decision trees (or weak learners) in various sub-samples of the data set during the training phase. Instead of relying on a single decision tree, RF considers the prediction by each decision tree and predicts the final output depending on the majority prediction by all the decision trees. It uses averaging to improve the predictive accuracy and control over-fitting. In classification problems, the final output of the RF will be the class predicted by most of the decision trees, whereas, in regression problems, the final output will be the average of the predicted outcome by all of the decision trees. In this study, we use RF for regression. We have used RF regressor implementations from \texttt{scikit-learn}, and more details can be found in \cite{35}.

\section{Optimisation Problem Formulation and Simulation Results}
\label{section3}
 
  We implement the proposed approach in the design of a bandgap reference circuit, a folded cascode op-amp, and a two-stage Miller compensated op-amp. The ML-assisted optimisation framework is implemented in Python, and the circuit simulations are carried out in \texttt{ngspice}. A specific ML model for a particular parameter is assigned through trial and error to identify the most promising model in each case. We use ML classification and regression models implementations from \texttt{scikit-learn} \cite{35}. The GridSearchCV in \texttt{scikit-learn} is used for tuning the parameters of every ML classification and regression model used. The parameters of an ML Model are optimized by cross-validated grid-search over a parameter grid. More details about GridSearchCV can be found in \cite{35}. A trained ML model can be used to predict the circuit characteristics without the need for retraining and predictions will be faster than subsequent circuit simulations. We also want to stress the re-usability of the trained ML model for multiple runs for validation.
  
  The simulation is carried out in a workstation with Intel\textsuperscript{\textregistered} Xeon\textsuperscript{\textregistered} Gold 6240R CPU @ 2.40\,GHz and 256\,GB RAM. Parallel processing has been implemented in the optimisation methodology to update each individual in the population, thereby reducing the algorithm's run time. For each circuit under study, 20 consecutive runs have been carried out with approaches SGA, MGA, MGA-MLSP, and MGA-MLSCP. For a fair comparison, the population size and the maximum number of generations are fixed as 20 and 200, respectively, in all the simulations. Out of the 20 runs, we discuss the best, worst, mean, and S.D. of the fitness function value and the average and median of spice calls per run for all the test cases considered and compare the results for different approaches.

\subsection{Bandgap Reference}

 \begin{figure}[!t]
    \centering
    \includegraphics[scale=0.8]{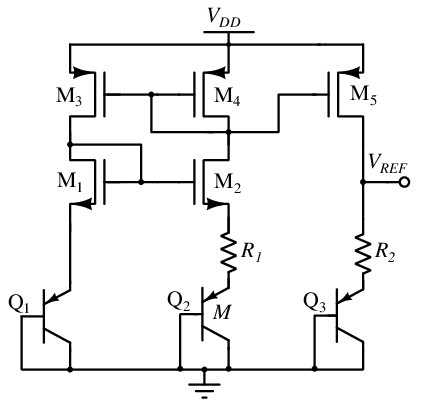}
    \caption{Schematic of bandgap reference.}
    \label{fig1}
  \end{figure}

      \begin{table}[!t]
    \begin{center}
    \caption{Design parameters and their ranges considered for BGR optimisation}
    \label{table1}
    \scalebox{0.95}{
      \begin{tabular}[!t]{ccc|ccc}
      \toprule      
      \textbf{Para-}&\textbf{LB}&\textbf{UB}&\textbf{Para-}&\textbf{LB}&\textbf{UB}\\
      \textbf{meter}&\textbf{}&\textbf{}&\textbf{meter}&\textbf{}&\textbf{}\\
       \midrule
        $W_{1,2}$&\SI{180}{\nano\meter}&\SI{50}{\micro\meter}&$R_2$&\SI{1}{\kilo\ohm}&\SI{150}{\kilo\ohm}\\
        $W_{3,4}$&\SI{180}{\nano\meter}&\SI{50}{\micro\meter}&$L_{1,2}$&\SI{180}{\nano\meter}&\SI{1}{\micro\meter}\\
        $W_{5}$&\SI{180}{\nano\meter}&\SI{50}{\micro\meter}&$L_{3,4}$&\SI{180}{\nano\meter}&\SI{1}{\micro\meter}\\
        $R_{1}$&\SI{500}{\ohm}&\SI{5}{\kilo\ohm}&$L_5$&\SI{180}{\nano\meter}&\SI{1}{\micro\meter}\\
        \bottomrule
      \end{tabular}} 
    \end{center}
  \end{table}
   Fig.~\ref{fig1} shows the circuit diagram of the BGR considered for optimisation. This circuit works on the principle of adding two voltages that vary in opposite directions with temperature, generating a reference voltage with zero overall temperature coefficient (TC). To frame the optimisation problem for the BGR, transistors M$_\text{1}$ and M$_\text{2}$, and M$_\text{3}$ and M$_\text{4}$ are assumed to be matched pairwise. The circuit specifications are considered as the constraints for this study. The saturation condition is also added as a constraint for the proper working of the circuit. The widths of the MOS transistors and the resistances in Fig.~\ref{fig1} are chosen as the decision variables. Since the main objective of a BGR is to produce a temperature-independent voltage, the TC of the BGR is considered as the fitness function.
  For the optimisation problem, we define the individual $\vb* x$ as 
  \begin{equation}
    \label{eq21}
    \vb*{x} = (W_{1,2}, W_{3,4}, W_{5}, R_1, R_2, L_{1,2}, L_{3,4}, L_5)
  \end{equation} 
  and the objective function, temperature coefficient (TC), as 
  \begin{equation}
    \label{eq22}
    TC = \frac{(V_{REF,\SI{125}{\celsius}}-V_{REF,\SI{-40}{\celsius}})\times\num{e6}}{V_{REF,\SI{27}{\celsius}}\times165}\,\,\SI{}{ppm/\celsius}.
  \end{equation}
    We then formulate the optimisation problem as follows:
\begin{equation*}
\begin{aligned}
& {\min}
&& TC \\
& \text{subject to}
& & \text{Power supply rejection ratio } (PSRR) \ge\SI{15}{\decibel} \\
&&& \text{Reference voltage variation } (\Delta V_{REF}) \le \SI{5}{\milli\volt}\\
&&& \text{Power dissipation }(P) \le\SI{600}{\micro\watt} \\
&&&  \SI{-40}{\celsius} \le \text{Temperature }(T) \le \SI{125}{\celsius} \\
&&& \text{Area }(A)\le \SI{500}{\micro\meter^{2}}\\
&&& 1 \le \text{Aspect ratio }(W/L) \le 100 \\
&&& \SI{180}{\nano\meter} \le \text{Length }(L) \le \SI{5}{\micro\meter} \\
&&&  \text{Noise} (S_n(f)) \le\SI{2}{\micro\volt\per\sqrt{Hz}}\text{ at } \SI{1}{\mega\hertz} \\
&&& \text{All transistors in saturation}.
\end{aligned}
\end{equation*}

    \begin{table}[!t]
    \begin{center}
    \caption{Performance of MLP classifiers, MLP(126, 64, 12) for saturation prediction for transistors in BGR optimisation.}
    \label{table2}
    \scalebox{0.95}{
      \begin{tabular}[!t]{cccc}
      \toprule        
      \textbf{Tranistor}&\textbf{Training}&\textbf{Testing}\\
      \textbf{}&\textbf{Time (s)}&\textbf{Accuracy}\\
        \midrule
        $M_1$@\SI{-40}{\celsius}&76.4&0.991\\
        $M_1$@\SI{125}{\celsius}&105.8&0.998\\
        $M_2$@\SI{-40}{\celsius}&101.7&0.975\\
        $M_2$@\SI{125}{\celsius}&96.4&0.987\\
        $M_3$@\SI{-40}{\celsius}&74.9&0.994\\
        $M_3$@\SI{125}{\celsius}&111.6&0.993\\
        $M_4$@\SI{-40}{\celsius}&80.8&0.999\\
        $M_4$@\SI{125}{\celsius}&69.1&0.998\\
        $M_5$@\SI{-40}{\celsius}&109.7&0.982\\
        $M_5$@\SI{125}{\celsius}&97.1&0.977\\
        \bottomrule
      \end{tabular}} 
    \end{center}
  \end{table}

     \begin{table}[!t]
    \begin{center}
    \caption{Performance of ML regression models for constraint prediction in BGR optimisation.}
    \label{table3}
    \scalebox{0.88}{
      \begin{tabular}[!t]{ccccc}
      \toprule      
      \textbf{Constraint}&\textbf{ML}&\textbf{Training}&\textbf{$\emph R^\emph 2$ }& \textbf{MAE}\\
      \textbf{Parameter}&\textbf{Model}&\textbf{Time (s)}&&\\
      \midrule
      $PSRR$&MLP&9.1&0.978&\num{8.64e-1}\\
      @ 1MHz (\si{\decibel})&(128,64,16)&&&\\
      $V_{REF}$&MLP&2.0&0.995&\num{1.6e-2}\\
      (\si{\volt})&(128,64,16)&&&\\
      $S_n(f)$&RF&4.8&0.927&\num{1.45e-7}\\
      (\si{\micro\volt\per\sqrt{Hz}})&(n=100)&&&\\
      \bottomrule
      \end{tabular}}
    \end{center}
    \end{table}
 
  The BGR circuit is designed in \SI{180}{\nano\meter} technology with a supply voltage of \SI{1.8}{\volt}. The typical-typical (TT) corner transistor models are used for the optimisation. The range for the design parameters considered in the optimisation problem is given in Table~\ref{table1}.

  MLP-based classifiers are trained to predict the saturation conditions. MLP classifiers are modeled at both extremities of the temperature range of the BGR; i.e., a total of ten MLP classifiers are built. ML regression models are trained for predicting the constraints of $PSRR$, $\Delta V_{REF}$, and $S_n(f)$. MLP-based regression models have been used for $PSRR$ and $V_{REF}$ prediction, and an RF regressor has been used for $S_n(f)$ prediction. A total of $13{,}000$ data points are generated using \texttt{ngspice} for the initial database. From which 80\% data points are used for training and 20\% data points are used for testing the ML models. The details of the ML classification and regression models are given in Table~\ref{table2} and Table~\ref{table3}, respectively. In Table~\ref{table3}, $R^2$, and $MAE$ correspond to R squared value and mean absolute error of the respective regression models. R-squared is a statistical measure which gives an idea about the fit of a regression model. MAE is the average of the absolute differences between predicted and actual values across the dataset. In the rest of the manuscript, the notation MLP ($a$, $b$, $c$) is used for MLP models where $a$, $b$, and $c$ represent the number of neurons in the first, second, and third layers, respectively. Similarly, the notation RF ($n = a$) is used for RF models, where $a$ represents the number of estimators.  

      \begin{table}[!t]
    \begin{center}
    \caption{Design parameters of the best solution obtained for BGR by SGA, MGA, MGA-MLSP, and MGA-MLSCP.}
    \label{table4}
    \scalebox{0.95}{
      \begin{tabular}[!t]{ccccc}
      \toprule       
      \textbf{Design}&\textbf{SGA}&\textbf{MGA}&\textbf{MGA-}&\textbf{MGA-}\\
      \textbf{Parameter}&\textbf{}&\textbf{}&\textbf{MLSP}&\textbf{MLSCP}\\
        \midrule
        $W_{1,2}$ (\si{\micro\meter}) &18.81&49.35&28.35&47.74\\
        $W_{3,4}$ (\si{\micro\meter}) &42.12&21.09&38.49&19.45\\
        $W_{5}$ (\si{\micro\meter}) &20.96&29.32&21.78&25.00\\
        $L_{1,2}$ (\si{\micro\meter})&0.81&1.43&1.26&1.76\\
        $L_{3,4}$ (\si{\micro\meter})&3.23&1.60&1.33&1.10\\
        $L_5$ (\si{\micro\meter})&4.39&2.35&0.88&0.78\\
        $R_1$ (\si{\kilo\ohm})&4.06&3.55&3.93&3.46\\
        $R_2$ (\si{\kilo\ohm})&108.86&36.06&41.63&16.86\\
        \bottomrule
      \end{tabular}} 
    \end{center}
  \end{table}

  \begin{table}[!t]
    \caption{BGR design specifications obtained for the best solution by SGA, MGA, MGA-MLSP, and MGA-MLSCP.}
    \label{table5}
    \begin{center}
    \scalebox{0.85}{
      \begin{tabular}{lcccccc}
        \toprule
        \textbf{Design}&{\bf Specs.}&\textbf{SGA}&\textbf{MGA}&\textbf{MGA-}&\textbf{MGA-}\\
        \textbf{Criteria}&{\bf}&\textbf{}&\textbf{}&\textbf{MLSP}&\textbf{MLSCP}\\
        \midrule
        $PSRR$ @1MHz&$\geq15$&16.64&17.11&16.63&16.90\\
        (\si{\decibel})&&&&&\\
        $P$ (\si{\milli\watt})&$\leq0.6$&0.063&0.091&0.079&0.120\\
        $A$ (\si{\micro\meter^{2}})&$\leq500$&394&277&192&231\\
        $S_n(f)$ @1MHz&$\leq2$&1.45&1.22&1.23&1.07\\
        (\si{\micro\volt\per\sqrt{Hz}})&&&&&\\
        $\Delta V_{REF}$ (\si{\milli\volt})&$\leq5$&0.408&0.234&0.205&0.196\\
        $V_{REF}$ (\si{\volt})&-&1.10&1.13&1.05&1.08\\
        \text{Saturation}&-&Met&Met&Met&Met\\
        \textbf{TC} (\si{ppm/\celsius})&min&\textbf{2.253}&\textbf{1.268}&\textbf{1.179}&\textbf{1.109}\\
        \bottomrule
      \end{tabular}} 
    \end{center}
  \end{table}

\begin{figure}[!t]
    \centering
    \includegraphics[scale=0.17]{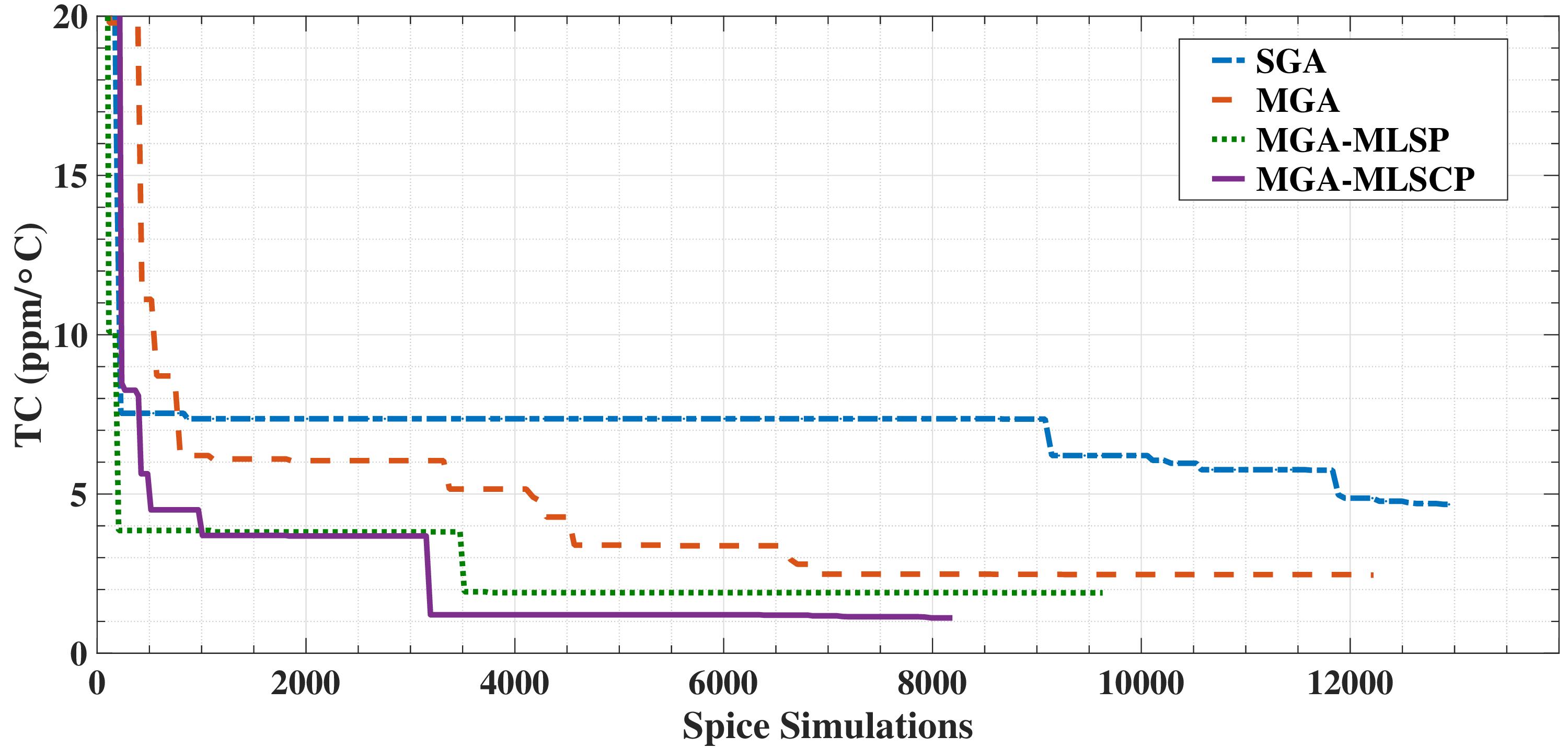}
    \caption{Comparison of convergence characteristics of SGA, MGA, MGA-MLSP, and MGA-MLSCP with respect to number of spice simulations for BGR optimisation. }
    \label{fig5}
  \end{figure}

  The design parameters of the best solution and the corresponding specifications obtained by SGA, MGA, MGA-MLSP, and MGA-MLSCP are given in Table~\ref{table4} and Table~\ref{table5}, respectively. The best, worst, mean, and S.D. of the fitness function value and the average number of spice calls per run for $20$ consecutive runs for all four algorithms are shown in Table~\ref{table18}. The MGA attains a better result for the TC with lesser S.D. and 39\% reduction in spice calls compared to SGA. MGA-MLSP and MGA-MLSCP attain better values for TC with further reduction in spice calls. MGA-MLSP and MGA-MLSCP converged to better values of \SI{1.179}{ppm/\celsius} and \SI{1.109}{ppm/\celsius} for TC with 51\% and 56\% reduction in the number of spice calls, respectively, when compareed to SGA. Fig.~\ref{fig5} shows the convergence characteristics of SGA, MGA, MGA-MLSP, and MGA-MLSCP with respect to spice simulations for a sample optimisation run for BGR. It is evident from the plot that MGA-MLSP and MGA-MLSCP are converging to optimal solution with fewer spice simulations than the other two algorithms.

\subsection{Folded Cascode Operational Amplifier}
 \begin{figure}[!t]
    \centering
    \includegraphics[scale=0.8]{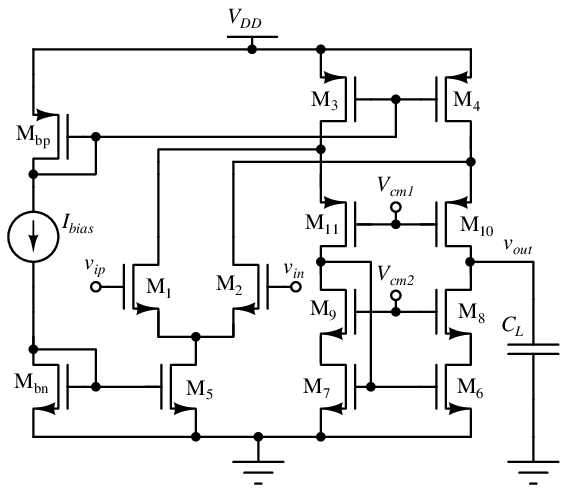}
    \caption{Schematic of a folded cascode op-amp.}
    \label{fig2}
  \end{figure}

  Fig.~\ref{fig2} shows the circuit diagram of the FCOA considered for optimisation. To frame the optimisation problem for the circuit, transistors M$_\text{1}$ and M$_\text{2}$, M$_\text{3}$, M$_\text{4}$, and M$_\text{bp}$, M$_\text{bn}$ and M$_\text{5}$, M$_\text{6}$ and M$_\text{7}$, M$_\text{8}$ and M$_\text{9}$, and M$_\text{10}$ and M$_\text{11}$ are assumed to be matched pairwise. The circuit specifications and the transistor saturation conditions are considered the constraints for this optimisation problem. After a detailed circuit study, the widths of the transistors and the bias current are chosen as the decision variables. The area of the FCOA is considered as the fitness function.
 The individual, $\vb* x$, and the objective function, $f(\vb* x)$, are given by:
  \begin{equation}
    \label{eq1}
    \vb*{x} = [W_{1,2}, W_{3,4,bp}, W_{bn,5}, W_{6,7}, W_{8,9}, W_{10,11}, I_{bias}]
  \end{equation} 
and 
  \begin{equation}
    \label{eq2}
    f(\vb*{x}) = \sum_{i=1}^{M}W_i\times L_i,
  \end{equation}
  respectively, where $M$ is the number of transistors in the circuit, and $W_i$ and $L_i$ are the width and length of the $i^\text{th}$ transistor. For the circuit in Fig.~\ref{fig2}, $M = 13$. We define the optimisation problem as follows.
\begin{equation*}
\begin{aligned}
& \underset{\vb*x}{\min}
& & f(\vb*x) \\
& \text{subject to}
& &\text{Voltage gain }(A_v) \ge\SI{40}{\decibel} \\
&&& \text{Power dissipation }(P) \le\SI{5}{\milli\watt} \\
&&& \text{Slew Rate }(SR)\ge\SI{20}{\volt\per\micro\s} \\
&&& \text{Unity gain bandwidth }(UGB) \ge\SI{40}{\mega\hertz} \\
&&& \text{Phase margin }(PM) \ge\SI{60}{\degree} \\
&&& \frac{4}{3} \le \text{Aspect ratio }(W/L) \le 300 \\
&&& \text{All transistors in saturation}.
\end{aligned}
\end{equation*}
  where ${\vb*x}$ and ${f(\vb*x)}$ are given by (\ref{eq1}) and (\ref{eq2}), respectively.
 
  The FCOA is designed in a \SI{180}{\nano\meter} technology. The supply voltage, $V_{DD}$, is taken as \SI{1.8}{\volt}. The transistor lengths are taken as \SI{180}{\nano\meter}. A value of \SI{5}{\pico\farad} \cite{22} is assumed for the load capacitance, $C_L$. The LB and UB for the decision variable, $I_{bias}$ is considered as \SI{1}{\micro\ampere} and \SI{1}{\milli\ampere}, respectively.

 MLP classifiers are trained to predict the saturation conditions of all $13$ transistors in FCOA. ML regression models are trained for predicting the constraints of $A_v$, $P$, $UGB$, $PM$, and $SR$. The MLP regression model is used for $A_v$ prediction. RF regression is used for all the other constraint predictions. From $20{,}000$ data points generated using \texttt{ngspice} for the initial database, 80\% data points are used for training, and 20\% data points are used for testing the ML models. The details of the ML classification and regression models are given in Table~\ref{table8} and Table~\ref{table9}, respectively.

      \begin{table}[!t]
    \begin{center}
    \caption{Performance of ML classifiers for saturation prediction of transistors for optimisation of FCOA.}
    \label{table8}
    \scalebox{0.95}{
      \begin{tabular}[!t]{ccccc}
      \toprule       
      \textbf{Trans-}&\textbf{ML}&\textbf{Training}&\textbf{Testing}\\
    \textbf{istor}&\textbf{Model}&\textbf{Time (s)}&\textbf{Accuracy}\\
        \midrule
        $M_1$&MLP(128,64,16)&130.6&0.997\\
        $M_2$&MLP(128,64,16)&183.2&0.998\\
        $M_3$&MLP(128,64,16)&246.7&0.996\\
        $M_4$&MLP(128,64,16)&250.9&0.996\\
        $M_5$&MLP(128,64,16)&256.2&0.990\\
        $M_6$&MLP(128,64,16)&259.3&0.990\\
        $M_7$&MLP(128,64,16)&180.8&0.993\\
        $M_8$&MLP(128,64,16)&229.2&0.990\\
        $M_9$&MLP(100,100)&223.7&0.991\\
        $M_{10}$&MLP(100,100)&117.8&0.993\\
        $M_{11}$&MLP(128,64,16)&116.2&0.993\\
        $M_{bn}$&MLP(128,64,16)&203.5&0.994\\
        $M_{bp}$&MLP(128,64,16)&191.0&0.994\\
        \bottomrule
      \end{tabular}} 
    \end{center}
  \end{table}

    \begin{table}[!t]
    \begin{center}
    \caption{Performance of ML regression models for constraint prediction in FCOA optimisation.}
    \label{table9}
    \scalebox{0.85}{
      \begin{tabular}[!t]{ccccc}
      \toprule       
      \textbf{Constraint}&\textbf{ML}&\textbf{Training}&\textbf{$\emph R^\emph 2$ }& \textbf{MAE}\\
      \textbf{Parameter}&\textbf{Model}&\textbf{Time (s)}&&\\
      \midrule
      $A_v$ (\si{\decibel})&MLP&147.1&0.900&\num{6.73e-1}\\
      &(128,64,20)&&&\\
      $P$ (\si{\milli\watt})&RF (n=500)&40.0&0.999&\num{1.605e-5}\\
      $UGB$ (\si{\mega\hertz})&RF(n=100)&11.4&0.978&\num{2.13e6}\\
      $PM$ $(^{\circ})$&RF(n=50)&3.8&0.942&\num{5.89e-1}\\
      $SR$ (\si{\volt\per\micro\second})&RF(n=500)&44.9&0.985&\num{1.01e6}\\
      \bottomrule
      \end{tabular}} 
    \end{center}
    \end{table}

    \begin{table}[!t]
    \begin{center}
    \caption{Optimum parameters obtained by SGA, MGA, MGA-MLSP, and MGA-MLSCP for FCOA optimisation.}
    \label{table10}
    \scalebox{0.95}{
    \begin{tabular}[]{lcccc}
        \toprule
        \textbf{Design}&{\bf SGA}&{\bf MGA}&{\bf MGA-}&{\bf MGA-}\\
        \textbf{Parameter}&{\bf}&{\bf}&{\bf MLSP}&{\bf MLSCP}\\
        \midrule
        $I_{bias}$ (\si{\micro\ampere})&288.5&269.7&263.4&258.9\\
        $W_{1,2}$ (\si{\nano\meter})&7195&8285&8551&8614\\
        $W_{3,4,bn}$ (\si{\nano\meter})&4349&3649&3554&3376\\
        $W_{5,bp}$ (\si{\nano\meter})&1615&1194&1176&1177\\
        $W_{6,7}$ (\si{\nano\meter})&1011&849&798&754\\
        $W_{8,9}$ (\si{\nano\meter})&539&392&417&374\\
        $W_{10,11}$ (\si{\nano\meter})&6401&6114&6017&6295\\
        \bottomrule
    \end{tabular}}
    \end{center}
    \end{table}

    \begin{table}[!t]
    \caption{Design specifications obtained for the area optimisation of FCOA by SGA, MGA, MGA-MLSP, and MGA-MLSCP.}
    \label{table11}
    \begin{center}
    \scalebox{0.85}{
      \begin{tabular}{lccccc}
        \toprule
        \textbf{Design}&{\bf Specs.}&{\bf SGA}&{\bf MGA}&{\bf MGA-}&{\bf MGA-} \\
        \textbf{Criteria}&{\bf}&{\bf}&{\bf}&{\bf MLSP}&{\bf MLSCP} \\
        \midrule       
        $A_v$ (\si{\decibel})&$\geq40$&40.47&40.51&40.96&41.01\\
        $UGB$ (\si{\mega\hertz})&$\geq40$&40.02&40.00&40.00&40.00\\
        $PM$ $(^{\circ})$&$\geq60$&89.97&89.92&89.87&89.89 \\
        $SR$ (\si{\volt\per\micro\second})&$\geq20$&22.19&21.91&21.02&21.53\\
        $P$ (\si{\milli\watt})&$\leq5$&1.379&1.281&1.251&1.230 \\
        Saturation&-&Met&Met&Met&Met\\
        $\textbf{\emph  A}$ (\si{\micro\meter^{2}})&min&\textbf{8.305}&\textbf{8.031}&\textbf{8.024}&\textbf{8.020}\\
        \bottomrule
    \end{tabular}}
    \end{center}
    \end{table}

        \begin{figure}[!t]
    \centering
    \includegraphics[scale=0.17]{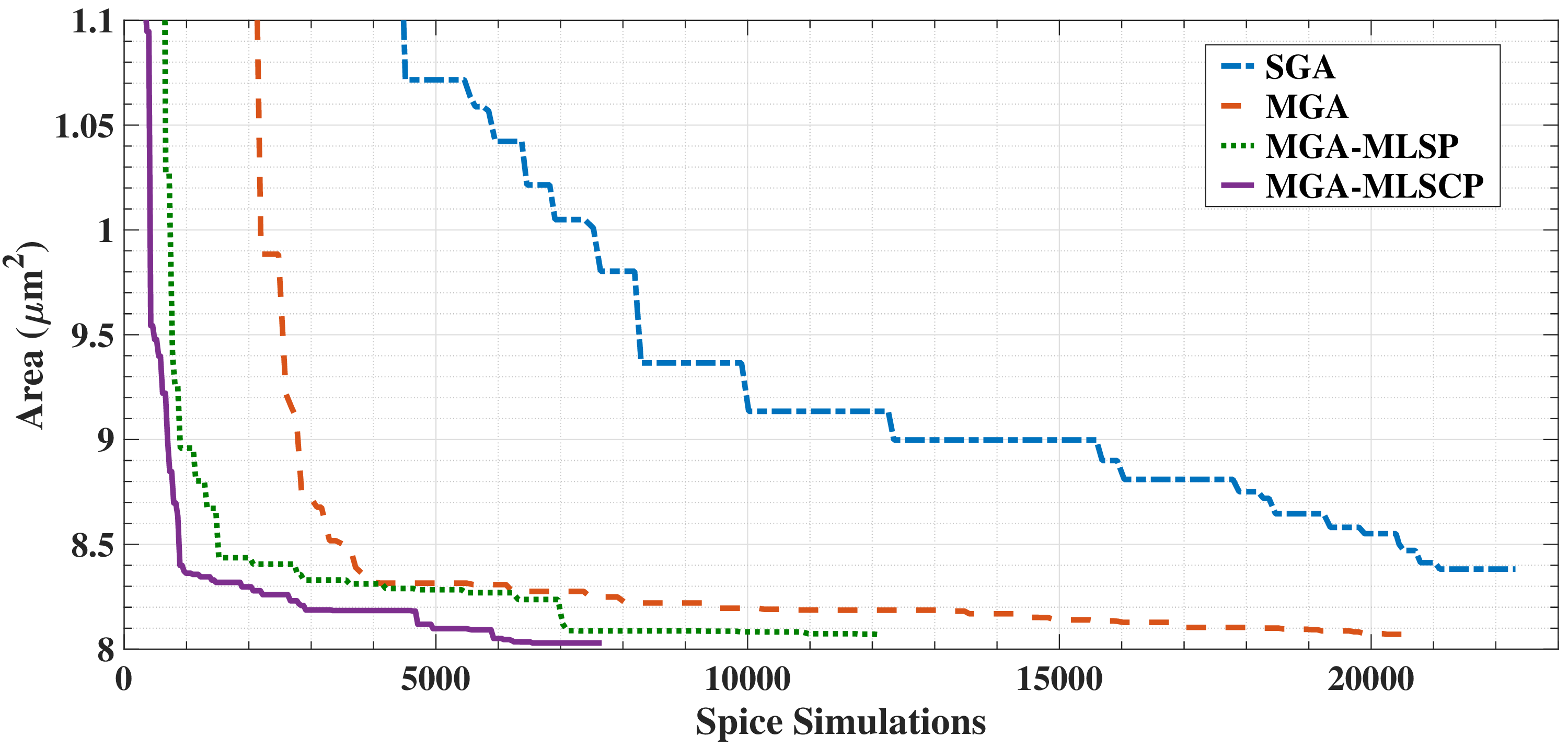}
    \caption{Comparison of convergence characteristics of SGA, MGA, MGA-MLSP, and MGA-MLSCP with respect to the number of spice simulations for FCOA optimisation.}
    \label{fig6}
  \end{figure}

  The design parameters of the best solution and the corresponding specifications obtained by SGA, MGA, MGA-MLSP, and MGA-MLSCP are summarised in Table~\ref{table10} and Table~\ref{table11}, respectively. Table~\ref{table18} compares $20$ consecutive runs for all four cases. The best, worst, mean, and S.D. of the fitness function value and the average number of spice calls per run for all the cases are presented. The MGA can attain a better result for the area with considerably less S.D.. MGA-MLSP and MGA-MLSCP attained better values for the area with a substantial reduction in spice calls. MGA-MLSP and MGA-MLSCP attained better values of \SI{8.024}{\micro\meter^2} and \SI{8.020}{\micro\meter^2} for the area with 43\% and 60\% reduction in the number of spice calls, respectively. Fig.~\ref{fig6} shows the convergence characteristics of SGA, MGA, MGA-MLSP, and MGA-MLSCP with respect to spice simulations for FCOA optimisation. It is evident that MGA-MLSP and MGA-MLSCP converge to the optimal solution with significantly fewer spice simulations when compared with SGA and MGA.

\subsection{Two-Stage Miller Compensated Operational Amplifier}

\begin{figure}[!t]
    \centering
    \includegraphics[width=0.8\linewidth]{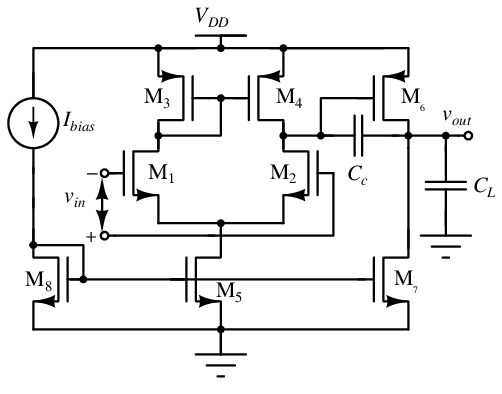}
    \caption{Schematic of a two-stage Miller compensated op-amp.}
    \label{fig3}
  \end{figure}

  A schematic of the two-stage Miller compensated op-amp is shown in Fig.~\ref{fig3}. It is assumed that the transistors M$_\text{1}$ and M$_\text{2}$, M$_\text{3}$ and M$_\text{4}$, and M$_\text{5}$ and M$_\text{8}$ are pairwise matched. As is common for op-amp optimisation, the circuit area is chosen as the fitness function. The widths and the bias current are chosen as the decision variables for the optimisation problem. All the design specifications are considered as constraints. Since all the transistors have to be in the saturation region of operation for the proper working of the circuit, the saturation condition is also added as a constraint. ICMR of \SI{0.6}{\volt} to \SI{1}{\volt} is also considered during the optimisation.
 
  The position vector, $\vb* x$, and the fitness function, $f(\vb* x)$, are
  \begin{equation}
    \label{eq3}
    \vb*{x} = [W_{1,2}, W_{3,4}, W_{5,8}, W_6, W_7, I_{bias}]
  \end{equation} 
and 
  \begin{equation}
    \label{eq4}
    f(\vb*{x}) = \sum_{i=1}^{M}W_i\times L_i,
  \end{equation}
  respectively, where $M$ is the total number of transistors in the circuit, and $W_i$ and $L_i$ are the width and length of the $i^\text{th}$ transistor. For the circuit in Fig.~\ref{fig3}, $M = 8$. 
  
  The optimisation problem is framed as:
\begin{equation*}
\begin{aligned}
& \underset{\vb*x}{\min}
&& f(\vb*x) \\
& \text{subject to}
& & \text{Voltage gain } (A_v) \ge\SI{20}{\decibel} \\
&&& \text{Power dissipation }(P) \le\SI{400}{\micro\watt} \\
&&& \text{Slew Rate }(SR) \ge\SI{100}{\volt\per\micro\s} \\
&&& \text{Cut-off frequency }(f_{3dB}) \ge\SI{10}{\mega\hertz} \\
&&& \text{Unity gain bandwidth }(UGB) \ge\SI{100}{\mega\hertz} \\
&&& \text{Phase margin }(PM)\ge\SI{60}{\degree} \\
&&& 2 \le \text{Aspect ratio }(W/L) \le 200 \\
&&& \text{Power spectral density }(S_n(f))\\
&&&\le\SI{60}{\nano\volt\per\sqrt{Hz}}\text{ at } \SI{1}{\mega\hertz} \\
&&& \text{All transistors in saturation}.
\end{aligned}
\end{equation*}
  where ${\vb*x}$ and ${f(\vb*x)}$ are given by (\ref{eq3}) and (\ref{eq4}), respectively.

          \begin{table}[!t]
    \begin{center}
    \caption{Performance of MLP classifiers (MLP(128, 64, 16)) for saturation prediction for transistors in TSMCOA optimisation.}
    \label{table14}
    \scalebox{0.95}{
      \begin{tabular}[!t]{cccc}
      \toprule      
      \textbf{Transistor}&\textbf{Training}&\textbf{Testing}\\
      \textbf{}&\textbf{Time (s)}&\textbf{Accuracy}\\
        \midrule
        $M_1@ICMR_{\min}$&168.7&0.994\\
        $M_1@ICMR_{\max}$&182.0&0.994\\
        $M_2@ICMR_{\min}$&179.1&0.994\\
        $M_2@ICMR_{\max}$&156.9&0.983\\
        $M_3@ICMR_{\min}$&126.1&0.995\\
        $M_3@ICMR_{\max}$&168.7&0.994\\
        $M_4@ICMR_{\min}$&129.4&0.994\\
        $M_4@ICMR_{\max}$&117.7&0.999\\
        $M_5@ICMR_{\min}$&173.5&0.996\\
        $M_5@ICMR_{\max}$&174.7&0.994\\
        $M_6@ICMR_{\min}$&189.8&0.991\\
        $M_6@ICMR_{\max}$&169.9&0.994\\
        $M_7@ICMR_{\min}$&145.7&0.984\\
        $M_7@ICMR_{\max}$&184.2&0.985\\
        $M_8@ICMR_{\min}$&90.85&0.981\\
        $M_8@ICMR_{\max}$&93.7&0.983\\
        \bottomrule
      \end{tabular}} 
    \end{center}
  \end{table}
      \begin{table}[!t]
    \begin{center}
    \caption{Performance of ML regression models for constraint prediction in TSMCOA optimisation.}
    \label{table15}
    \scalebox{0.85}{
      \begin{tabular}[!t]{ccccc}
      \toprule     
      \textbf{Constraint}&\textbf{ML}&\textbf{Time}&\textbf{$\emph R^\emph 2$ }& \textbf{MAE}\\
      \textbf{Parameter}&\textbf{Model}&\textbf{(s)}&&\\
      \midrule
      $A_v@ICMR_{\min}$&MLP&18.1&0.814&1.842\\
      (\si{\decibel})&(50,50,50,50,50)&&&\\
      $A_v@ICMR_{\max}$&MLP&49.3&0.827&1.230\\
      (\si{\decibel})&(50,50,50,50)&&&\\
      $UGB@ICMR_{\min}$&MLP&32.6&0.980&0.816\\
      (\si{\mega\hertz})&(256,256)&&&\\
      $UGB@ICMR_{\max}$&MLP&9.1&0.980&0.816\\
      (\si{\mega\hertz})&(100,50,25,5)&&&\\
      $S_n(f)@ICMR_{\min}$&RF&9.1&0.980&0.816\\
      (\si{\nano\volt\per\sqrt{Hz}})&(n=2000)&&&\\
      $S_n(f)@ICMR_{\max}$&RF&9.1&0.980&0.816\\
      (\si{\nano\volt\per\sqrt{Hz}})&(n=1000)&&&\\
      $PM$ $(^{\circ})$&RF(n=100)&2.0&0.995&0.015\\
      $f_{3dB}$&MLP&4.8&0.989&5.25\\
      (\si{\mega\hertz})&(50,50,50,50,50)&&&\\
      \bottomrule
      \end{tabular}} 
    \end{center}
    \end{table}

  The two-stage op-amp \cite{20} is designed in a \SI{65}{\nano\meter} technology. The length of the transistors is fixed to the minimum of \SI{60}{\nano\meter}, with the supply voltage as \SI{1.1}{\volt}. The other specifications of voltage gain, power dissipation, slew rate, etc., are fixed accordingly. For example, a gain specification of \SI{20}{\decibel} for a transistor length of \SI{60}{\nano\meter} is considered here. The value of load capacitor, $C_L$, is taken as \SI{200}{\femto\farad}. The value of $C_c$ is taken as 0.3 times $C_L$ \cite{1}. The lower bound and upper bound for the decision variable, $I_{bias}$, is considered as \SI{1}{\micro\ampere} and \SI{100}{\micro\ampere}, respectively.

    \begin{table}[!t]
    \begin{center}
    \caption{Optimum parameters obtained by SGA, MGA, MGA-MLSP, and MGA-MLSCP for TSMCOA optimisation.}
    \label{table16}
    \scalebox{0.95}{
    \begin{tabular}[]{lcccc}
        \toprule
        \textbf{Design}&{\bf SGA}&{\bf MGA}&{\bf MGA-}&{\bf MGA-}\\
         \textbf{Parameter}&{\bf }&{\bf }&{\bf MLSP}&{\bf MLSCP}\\
        \midrule
        $I_{bias}$ (\si{\micro\ampere})&32.3&28.9&27.6&28.0\\
        $W_{1,2}$ (\si{\nano\meter})&287&259&249&252\\
        $W_{3,4}$ (\si{\nano\meter})&724&799&821&815\\
        $W_{5,8}$ (\si{\nano\meter})&139&124&130&119\\
        $W_6$ (\si{\nano\meter})&1531&1119&1082&1122\\
        $W_7$ (\si{\nano\meter})&195&187&193&181\\
        \bottomrule
        \end{tabular}}
        \end{center}
    \end{table}

    \begin{table}[!t]
    \caption{Design Specifications obtained for the area optimisation of TSMCOA by SGA, MGA, MGA-MLSP, and MGA-MLSCP.}
    \label{table17}
    \begin{center}
    \scalebox{0.85}{
      \begin{tabular}{lccccc}
        \toprule
        \textbf{Design}&{\bf Specs.}&{\bf SGA}&{\bf MGA}&{\bf MGA-}&{\bf MGA-} \\
        \textbf{Criteria}&{\bf}&{\bf}&{\bf }&{\bf MLSP}&{\bf MLSCP} \\
        \midrule
        $A_v$ (\si{\decibel})&$\geq20$&24.6&21.9&21.5&21.87\\
        $f_{3dB}$ (\si{\mega\hertz})&$\geq10$&11.72&13.29&13.18&12.99\\
        $UGB$ (\si{\mega\hertz})&$\geq100$&187.7&156.3&149.1&152.5\\
        $PM$ $(^{\circ})$&$\geq60$&60.0&60.0&60.1&60.1 \\
        $SR$ (\si{\volt\per\micro\second})&$\geq100$&265&269&268&264 \\
        $S_n(f)$@1MHz&$\leq60$&52.94&53.29&53.41&53.36\\
        (\si{\nano\volt\per\sqrt{Hz}})&&&&&\\
        Saturation&-&Met&Met&Met&Met\\
        $\textbf{\emph A}$ (\si{\micro\meter^{2}})&$\leq1$&\textbf{0.2416}&\textbf{0.2206}&\textbf{0.2205}&\textbf{0.2205}\\
        \bottomrule
        \end{tabular}}
        \end{center}
    \end{table}

  $16$ MLP classifiers are trained for predicting the saturation conditions of all 8 transistors at both extremes of the ICMR in TSMCOA. MLP regression models are trained for predicting the constraints of $A_v$, $UGB$, and $f_{3dB}$. RF regression models are used for $PM$ and $S_n(f)$ prediction. A total of $25{,}000$ data points are generated using \texttt{ngspice} for the initial database, where 80\% data points are used for training, and 20\% data points are used for testing the ML models. The details of the ML classification and regression models are given in Table~\ref{table14} and Table~\ref{table15}, respectively.

  The design parameters of the best solution and the corresponding specifications obtained by SGA, MGA, MGA-MLSP, and MGA-MLSCP are given in Table~\ref{table16} and Table~\ref{table17}, respectively. Table~\ref{table18} compares $20$ consecutive runs for all four cases. The best, worst, mean, and S.D. of the fitness function value and the average and the median of spice simulations invoked per run for all the cases are presented. The MGA was able to attain a better result for the area with considerably less S.D. and with 43\% reduction in spice calls than the SGA. MGA-MLSP and MGA-MLSCP attained a better value of \SI{0.2205}{\micro\meter^2} for the area with 76\% and 84\% reduction in the number of spice calls, respectively. 

\begin{figure}[!t]
    \centering
    \includegraphics[scale=0.17]{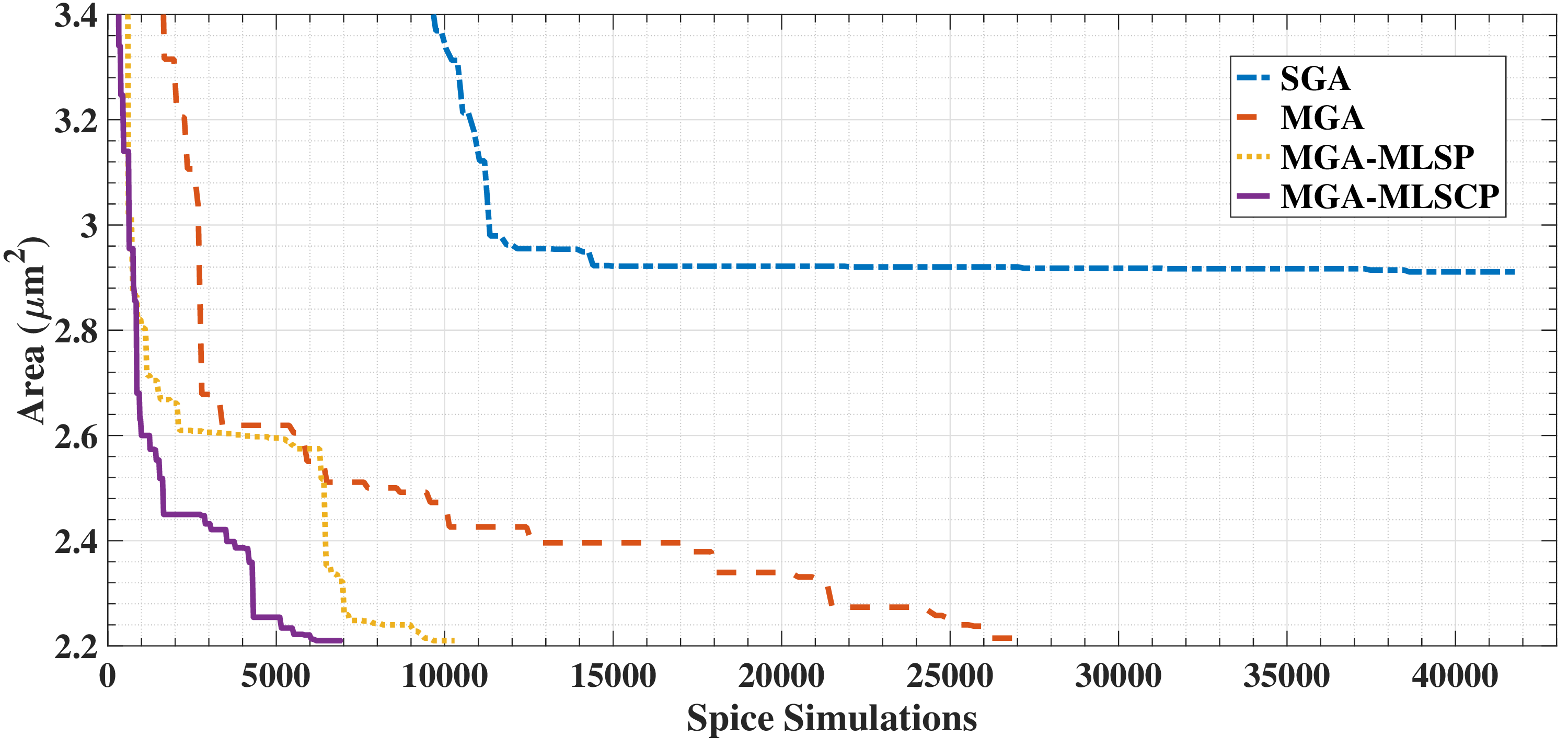}
    \caption{Comparison of convergence characteristics of SGA, MGA, MGA-MLSP, and MGA-MLSCP with respect to the number of spice simulations for TSMCOA optimisation.}
    \label{fig7}
  \end{figure}

  Fig.~\ref{fig7} shows the convergence characteristics of SGA, MGA, MGA-MLSP, and MGA-MLSCP with respect to spice simulations for TSMCOA optimisation. Since ICMR is also considered during the optimisation of TSMCOA, the specifications are more stringent. The optimised design needs to satisfy all the constraints in this range. This is reflected by the considerably higher number of spice simulations invoked by the SGA and MGA during optimisation when compared with the other two test cases. MGA-MLSP and MGA-MLSCP converged to the optimal solution with a significant reduction in spice calls, which is evident from Fig.~\ref{fig7}. This shows that the proposed ML-driven algorithms perform better when the circuit specifications are more stringent. Although the initial database for training the ML models is seemingly large for MGA-MLSP and MGA-MLSCP, it is compensated by the substantial reduction in the number of spice calls undertaken in subsequent runs by the proposed two algorithms in comparison with MGA and SGA. This is true for all the test cases considered.

    \begin{table}[!t]
    \begin{center}
    \caption{Performance of algorithms SGA, MGA, MGA-MLSP, and MGA-MLSCP, for BGR, FCOA, and TSMCOA optimisation.}
    \label{table18}
    \scalebox{0.8}{
      \begin{tabular}[!t]{ccccccc}
      \toprule
      \multicolumn{7}{c}{\textbf{Bandgap Reference} (\textbf{BGR})}\\
      \midrule
      \multirow{3}{*}{\textbf{Algorithm}}&\multicolumn{4}{c}{\textbf{Optimal TC}} & \multicolumn{2}{c}{\textbf{Spice}}\\
      &\multicolumn{4}{c}{(\si{ppm/\celsius})} & \multicolumn{2}{c}{\textbf{Simulations }(\#)}\\
      \cmidrule(lr){2-5}\cmidrule(lr){6-7}
    \textbf{}& \textbf{Best}&\textbf{Worst}&\textbf{Mean}&\textbf{S.D.}&\textbf{Mean}&\textbf{Median}\\
        \midrule
        SGA &2.253&9.030&4.570&2.421&18595& 13614\\
        MGA &1.268&3.378&2.102&0.554&11318&11084\\
        MGA-MLSP&1.179&3.052&1.941&0.456&9124&8942\\
        MGA-MLSCP &1.109&2.257&1.593&0.294&8178&8154\\
        \midrule
        \multicolumn{7}{c}{\textbf{Folded cascode op-amp} (\textbf{FCOA})}\\
        \midrule
      \multirow{3}{*}{\textbf{Algorithm}}&\multicolumn{4}{c}{\textbf{Optimal Area}} & \multicolumn{2}{c}{\textbf{Spice}}\\
      &\multicolumn{4}{c}{(\si{\micro\meter^{2}})} & \multicolumn{2}{c}{\textbf{Simulations }(\#)}\\
      \cmidrule(lr){2-5}\cmidrule(lr){6-7}
    \textbf{}& \textbf{Best}&\textbf{Worst}&\textbf{Mean}&\textbf{S.D.}&\textbf{Mean}&\textbf{Median}\\
    \midrule
        SGA &8.305&10.520&8.899&0.660&21518&21084 \\
        MGA &8.031&8.276&8.086&0.069&20936&20926\\
        MGA-MLSP&8.024&8.128&8.065&0.041&12176&12133\\
        MGA-MLSCP &8.020&8.088&8.043&0.022&8639&8755\\
        \midrule
        \multicolumn{7}{c}{\textbf{Two-stage Miller compensated op-amp} (\textbf{TSMCOA})}\\
        \midrule
      \multirow{3}{*}{\textbf{Algorithm}}&\multicolumn{4}{c}{\textbf{Optimal Area}}
      & \multicolumn{2}{c}{\textbf{Spice}}\\
      &\multicolumn{4}{c}{(\si{\micro\meter^{2}})} & \multicolumn{2}{c}{\textbf{Simulations }(\#)}\\
      \cmidrule(lr){2-5}\cmidrule(lr){6-7}
    \textbf{}& \textbf{Best}&\textbf{Worst}&\textbf{Mean}&\textbf{S.D.}&\textbf{Mean}&\textbf{Median}\\
    \midrule
        SGA &0.2416&0.3128&0.2564&0.0220&43521&42836\\
        MGA &0.2206&0.2348&0.2258&0.0052&24908&24405\\
        MGA-MLSP&0.2205&0.2264&0.2227&0.0021&10265&10255\\
        MGA-MLSCP &0.2205&0.2234&0.2218&0.0010&6994&7203\\
        \bottomrule
      \end{tabular} }
    \end{center}
  \end{table}

  We have considered the single- and multi-objective optimisation for TSMCOA reported in \cite{37} for comparison with the state-of-the-art. In \cite{37}, a neural network-based high-efficiency optimisation method is utilised for single- and multi-objective optimisations of a TSMCOA. All the constraints and specifications used in \cite{37} are considered for a fair comparison and the objective function for single-objective optimisation is chosen as the area. Area and power are selected as the objective functions to be optimised in the case of multi-objective optimisation, and the fitness function is expressed as: 
  \begin{equation}
    \label{eq7}
    f(\vb*{x}) = \alpha \sum_{i=1}^{M}W_i\times L_i + \beta P,
  \end{equation}
  where $\alpha$ and $\beta$ are the weights of area and power, respectively. The results of the single-objective and multi-objective optimisation of TSMCOA in comparison with \cite{37} are presented in Table.~\ref{table19} and Table.~\ref{table20}, respectively. In the area optimisation, the proposed method is able to achieve a 38\% reduction in area compared to \cite{37}. In multi-objective optimisation, the results show that MGA-MLSCP is able to attain a better value for area with reduced power than \cite{37}. There is a 21\% and 16\% reduction in area and power, respectively. 

            \begin{table}[!t]
        \caption{Comparison of area optimisation of TSMCOA using MGA-MLSCP with ANN-PSO \cite{37}.}
    \label{table19}
    \begin{center}
    \scalebox{0.95}{
      \begin{tabular}{lccc}
        \toprule
        \textbf{Design}&\textbf{Specs.}&\textbf{ANN-PSO}&\textbf{This} \\
       \textbf{Criteria}&\textbf{}&\textbf{\cite{37}}&\textbf{Work} \\
        \midrule
        $A_v$ (\si{\decibel})&$\geq20$&20.18&20.49\\
        $UGB$ (\si{\mega\hertz})&$\geq100$&106.4&108.49\\
        $PM$ $(^{\circ})$&$\geq60$&60.09&61.58\\
        $SR$ (\si{\volt\per\micro\second})&$\geq100$&176&160\\
        $CMRR$ (\si{\decibel})&-&30.3&25.42\\
        $PSRR$ (\si{\decibel})&-&15.64&27.32\\
        $V_{DD}$ (\si{\volt})&-&1.1&1.1\\
        $V_{SS}$ (\si{\volt})&-&0&0\\
        $P$ (\si{\micro\watt})&$\leq150$&83.2&90.96\\
        $\textbf{\emph A}$ (\si{\micro\meter^{2}})&$\leq1$&\textbf{0.121}&\textbf{0.075}\\
        \bottomrule
        \end{tabular}}
        \end{center}
    \end{table}

      \begin{table}[!t]
    \caption{Comparison of area and power optimisation of TSMCOA using MGA-MLSCP with ANN-PSO \cite{37}.}
    \label{table20}
    \begin{center}
    \scalebox{0.95}{
      \begin{tabular}{lccc}
        \toprule
        \textbf{Design}&\textbf{Specs.}&\textbf{ANN-PSO}&\textbf{This} \\
        \textbf{Criteria}&\textbf{}&\textbf{\cite{37}}&\textbf{Work} \\
        \midrule
        $A_v$ (\si{\decibel})&$\geq20$&21.36&20.29\\
        $UGB$ (\si{\mega\hertz})&$\geq100$&104.7&108.98\\
        $PM$ $(^{\circ})$&$\geq60$&60.68&60.27\\
        $SR$ (\si{\volt\per\micro\second})&$\geq100$&164&153\\
        $S_n(f)$ @1MHz&$\leq80$&79.22&77.09\\
         (\si{\nano\volt\per\sqrt{Hz}})&&&\\
        $S_n(f)$ @10MHz&-&35.03&26.27\\
        (\si{\nano\volt\per\sqrt{Hz}})&&&\\
        $CMRR$ (\si{\decibel})&-&32.08&32.22\\
        $PSRR$ (\si{\decibel})&-&18.43&30.50\\
        Settling time &-&9.48&8.57\\ 
        with 2\% tol.(\si{\nano\second})&&&\\
        Settling time &-&7.899&5.75\\
        with 5\% tol.(\si{\nano\second})&&&\\
        $V_{DD}$ (\si{\volt})&-&1.1&1.1\\
        $V_{SS}$ (\si{\volt})&-&0&0\\
        $\textbf{\emph P}$ (\si{\micro\watt})&$\leq150$&\textbf{72.09}&\textbf{60.45}\\
        $\textbf{\emph A}$ (\si{\micro\meter^{2}})&$\leq1$&\textbf{0.1371}&\textbf{0.1078}\\
        \bottomrule
        \end{tabular}}
        \end{center}
    \end{table}

\section{Conclusion}
\label{section4}
 
We present a machine learning machine learning-driven optimisation framework using genetic algorithm to design optimal analog circuits. Combining spice simulations with ML predictions guides genetic algorithm to optimum solutions. 
Since it is vital to ensure the right operating region of the transistors for the proper working of the analog circuits, we have used ML classification models to predict the operating region of the transistors in the analog circuit. ML regression models are used to predict the other circuit constraint specifications like power, gain, phase margin, etc. The simulation results show that the proposed framework can attain better solutions with higher optimisation efficiency while invoking fewer spice simulations. We validated the proposed methodology to optimise a bandgap reference, a folded cascade operational amplifier, and a two-stage Miller-compensated operational amplifier. The results show that the proposed approach attains better solutions for the fitness functions with a reduction of 53\%, 56\%, and 83\% spice calls in the three test cases considered when compared to the standard genetic algorithm and its modifications. Comparison with the state-of-the-art highlights the superiority of the proposed framework in attaining a better optimal value. We plan to explore employing this framework in more intricate analog circuits, such as phase-locked loops and comparators in the future.  

\section*{Acknowledgements}
\label{section5}
 We thank Dr. Rudra Narayan Roy, School of Mechanical Sciences, IIT Goa for providing access to a high-speed computational facility.

 \bibliographystyle{elsarticle-num} 
 \bibliography{references}

\begin{thebibliography}{10}
\expandafter\ifx\csname url\endcsname\relax
  \def\url#1{\texttt{#1}}\fi
\expandafter\ifx\csname urlprefix\endcsname\relax\def\urlprefix{URL }\fi
\expandafter\ifx\csname href\endcsname\relax
  \def\href#1#2{#2} \def\path#1{#1}\fi

\bibitem{1}
B.~Razavi, {Design of Analog CMOS Integrated Circuits}, 2nd Edition, McGraw-Hill Education, New York, USA, 2017.

\bibitem{2}
A.~A. Youssef, B.~Murmann, H.~Omran, {Analog IC Design using Precomputed Lookup Tables: Challenges and Solutions}, IEEE Access 8 (2020) 134640--134652.

\bibitem{3}
P.~Walker, J.~P. Ochoa-Ricoux, A.~Abusleme, {Slice-based Analog Design}, IEEE Access 9 (2021) 148164--148183.

\bibitem{4}
S.~Yin, R.~Wang, J.~Zhang, Y.~Wang, {Asynchronous Parallel Expected Improvement Matrix-Based Constrained Multi-Objective Optimization for Analog Circuit Sizing}, IEEE Transactions on Circuits and Systems II: Express Briefs 69~(9) (2022) 3869--3873.

\bibitem{48}
J.~Tao, Y.~Su, D.~Zhou, X.~Zeng, X.~Li, {Graph-Constrained Sparse Performance Modeling for Analog Circuit Optimization via SDP Relaxation}, IEEE Transactions on Computer-Aided Design of Integrated Circuits and Systems 38~(8) (2019) 1385--1398.

\bibitem{105}
E.~Afacan, Inversion coefficient optimization based analog/rf circuit design automation, Microelectronics Journal 83 (2019) 86--93.

\bibitem{106}
N.~Sabry, H.~Omran, M.~Dessouky, Systematic design and optimization of operational transconductance amplifier using gm/id design methodology, Microelectronics Journal 75 (2018) 87--96.

\bibitem{5}
S.~P. Boyd, S.~J. Kim, {Geometric Programming for Circuit Optimization}, in: Proc. of the 2005 Int. Symp. on Physical Design, ISPD '05, Association for Computing Machinery, NY, USA, 2005, p. 44–46.

\bibitem{6}
J.~Koza, F.~Bennett, D.~Andre, M.~Keane, F.~Dunlap, {Automated Synthesis of Analog Electrical Circuits by means Of Genetic Programming}, IEEE Transactions on Evolutionary Computation 1~(2) (1997) 109--128.

\bibitem{7}
E.~Hjalmarson, R.~Hagglund, L.~Wanhammar, {An Equation-Based Optimization Approach for Analog Circuit Design}, in: International Symposium on Signals, Circuits and Systems, Vol.~1, 2003, pp. 77--80.

\bibitem{8}
G.~Alpaydin, S.~Balkir, G.~Dundar, {An Evolutionary Approach to Automatic Synthesis of High-Performance Analog Integrated Circuits}, IEEE Transactions on Evolutionary Computing 7~(3) (2003) 240--252.

\bibitem{51}
A.~Patanè, A.~Santoro, P.~Conca, G.~Carapezza, A.~L. Magna, V.~Romano, G.~Nicosia, {Multi-Objective Optimization and Analysis for the Design Space Exploration of Analog Circuits and Solar Cells}, Engineering Applications of Artificial Intelligence 62 (2017) 373--383.

\bibitem{107}
M.~Sabry, I.~Nashaat, H.~Omran, Automated design and optimization flow for fully-differential switched capacitor amplifiers using recycling folded cascode ota, Microelectronics Journal 101 (2020) 104814.

\bibitem{108}
E.~Afacan, G.~Berkol, G.~Dundar, F.~Başkaya, An analog circuit synthesis tool based on efficient and reliable yield estimation, Microelectronics Journal 54 (05 2016).

\bibitem{109}
A.~Andrade, A.~Petraglia, C.~Soares, A constrained optimization approach for accurate and area efficient bandgap reference design, Microelectronics Journal 65 (2017) 72--77.

\bibitem{113}
J.~Li, Y.~Li, Y.~Zeng, {Robust Circuit Optimization under PVT Variations via Weight Optimization Problem Reformulation}, Expert Systems with Applications 248 (2024) 123301.

\bibitem{31}
W.~Lyu, P.~Xue, F.~Yang, C.~Yan, Z.~Hong, X.~Zeng, D.~Zhou, {An Efficient Bayesian Optimization Approach for Automated Optimization of Analog Circuits}, IEEE Transactions on Circuits and Systems I: Regular Papers 65~(6) (2018) 1954--1967.

\bibitem{9}
Y.~Yang, H.~Zhu, Z.~Bi, C.~Yan, D.~Zhou, Y.~Su, X.~Zeng, {Smart-MSP: A Self-Adaptive Multiple Starting Point Optimization Approach for Analog Circuit Synthesis}, IEEE Transactions on Computer-Aided Design of Integrated Circuits and Systems 37~(3) (2018) 531--544.

\bibitem{10}
G.~I. Tombak, S.~N. Güzelhan, E.~Afacan, G.~Dündar, {Simulated Annealing Assisted NSGA-III-based Multi-Objective Analog IC Sizing Tool}, Integration 85 (2022) 48--56.

\bibitem{49}
R.~Martins, N.~Lourenço, R.~Póvoa, N.~Horta, {Shortening the Gap between Pre- and Post-Layout Analog IC Performance by Reducing the LDE-Induced Variations with Multi-Objective Simulated Quantum Annealing}, Engineering Applications of Artificial Intelligence 98 (2021) 104102.

\bibitem{11}
S.~Zhang, F.~Yang, C.~Yan, D.~Zhou, X.~Zeng, {An Efficient Batch-Constrained Bayesian Optimization Approach for Analog Circuit Synthesis via Multiobjective Acquisition Ensemble}, IEEE Transactions on Computer-Aided Design of Integrated Circuits and Systems 41~(1) (2022) 1--14.

\bibitem{111}
B.~He, S.~Zhang, Y.~Wang, T.~Gao, F.~Yang, C.~Yan, D.~Zhou, Z.~Bi, X.~Zeng, {A Batched Bayesian Optimization Approach for Analog Circuit Synthesis via Multi-Fidelity Modeling}, IEEE Transactions on Computer-Aided Design of Integrated Circuits and Systems 42~(2) (2023) 347--359.

\bibitem{12}
M.~Fayazi, Z.~Colter, E.~Afshari, R.~Dreslinski, {Applications of Artificial Intelligence on the Modeling and Optimization for Analog and Mixed-Signal Circuits: A Review}, IEEE Transactions on Circuits and Systems I: Regular Papers 68~(6) (2021) 2418--2431.

\bibitem{13}
M.~Dehbashian, M.~Maymandi-Nejad, {An Enhanced Optimization Kernel for Analog IC Design Automation using the Shrinking Circles Technique}, Engineering Applications of Artificial Intelligence 58 (2017) 62--78.

\bibitem{14}
B.~Liu, D.~Zhao, P.~Reynaert, G.~G.~E. Gielen, {Synthesis of Integrated Passive Components for High-Frequency RF ICs Based on Evolutionary Computation and Machine Learning Techniques}, IEEE Transactions on Computer-Aided Design of Integrated Circuits and Systems 30~(10) (2011) 1458--1468.

\bibitem{112}
J.~Li, Y.~Zeng, H.~Zhi, J.~Yang, W.~Shan, Y.~Li, Y.~Li, {Knowledge Transfer Framework for PVT Robustness in Analog Integrated Circuits}, IEEE Transactions on Circuits and Systems I: Regular Papers 71~(5) (2024) 2017--2030.

\bibitem{15}
B.~Liu, Y.~Wang, Z.~Yu, L.~Liu, M.~Li, Z.~Wang, J.~Lu, F.~Fernandez, {Analog Circuit Optimization System based on Hybrid Evolutionary Algorithms}, Integration, the VLSI Journal 42 (2009) 137--148.

\bibitem{16}
G.~Nicosia, S.~Rinaudo, E.~Sciacca, {An Evolutionary Algorithm-based Approach to Robust Analog Circuit Design using Constrained Multi-Objective Optimization}, Knowledge-Based Systems 21~(3) (2008) 175--183, aI 2007.

\bibitem{17}
S.~Ghosh, B.~P. De, R.~Kar, D.~Mandal, A.~K. Mal, {Optimal Design of Complementary Metal-Oxide-Semiconductor Analogue Circuits: An Evolutionary Approach}, Computers \& Electrical Engineering 80 (2019) 106485.

\bibitem{50}
Žiga Rojec, Árpád Bűrmen, I.~Fajfar, {Analog Circuit Topology Synthesis by means of Evolutionary Computation}, Engineering Applications of Artificial Intelligence 80 (2019) 48--65.

\bibitem{18}
C.~Vişan, O.~Pascu, M.~Stănescu, E.~D. Şandru, C.~Diaconu, A.~Buzo, G.~Pelz, H.~Cucu, {Automated Circuit Sizing with Multi-Objective Optimization based on DE and Bayesian Inference}, Knowledge-Based Systems 258 (2022) 109987.

\bibitem{19}
R.~A. d.~L.~Moreto, C.~E. Thomaz, S.~P. Gimenez, {A Customized Genetic Algorithm with In-Loop Robustness Analyses to Boost the Optimization Process of Analog CMOS ICs}, Microelectronics Journal 92 (2019) 104595.

\bibitem{110}
R.~A. de~Lima~Moreto, C.~E. Thomaz, S.~P. Gimenez, A customized genetic algorithm with in-loop robustness analyses to boost the optimization process of analog cmos ics, Microelectronics Journal 92 (2019) 104595.

\bibitem{34}
R.~Rashid, G.~Raghunath, V.~Badugu, N.~Nambath, {Performance Evaluation of Evolutionary Algorithms for Analog Integrated Circuit Design Optimisation}, Microelectronics Journal 141 (2023) 105983.

\bibitem{103}
A.~Lberni, M.~A. Marktani, A.~Ahaitouf, A.~Ahaitouf, {Efficient Butterfly Inspired Optimization Algorithm for Analog Circuits Design}, Microelectronics Journal 113 (2021) 105078.

\bibitem{20}
R.~Rashid, N.~Nambath, {Area Optimisation of Two Stage Miller Compensated Op-Amp in 65 nm Using Hybrid PSO}, IEEE Transactions on Circuits and Systems II: Express Briefs 69~(1) (2022) 199--203.

\bibitem{21}
R.~Phelps, M.~Krasnicki, R.~A. Rutenbar, L.~R. Carley, J.~R. Hellums, {Anaconda: Simulation-Based Synthesis of Analog Circuits via Stochastic Pattern Search}, Trans. Comp.-Aided Des. Integ. Cir. Sys. 19~(6) (2006) 703–717.

\bibitem{22}
F.~V. Liu, B.and~Fern\'{a}ndez, G.~Gielen, R.~Castro-L\'{o}pez, E.~Roca, {A Memetic Approach to the Automatic Design of High-Performance Analog Integrated Circuits}, ACM Trans. Des. Autom. Electron. Syst. 14~(3) (jun 2009).

\bibitem{23}
B.~Liu, N.~Deferm, D.~Zhao, P.~Reynaert, G.~Gielen, {An Efficient High-Frequency Linear RF Amplifier Synthesis Method based on Evolutionary Computation and Machine Learning Techniques}, IEEE Transactions on Computer-aided Design of Integrated Circuits and Systems 31 (01 2011).

\bibitem{24}
G.~Wolfe, R.~Vemuri, {Extraction and Use of Neural Network Models in Automated Synthesis of Operational Amplifiers}, IEEE Transactions on Computer-Aided Design of Integrated Circuits and Systems 22~(2) (2003) 198--212.

\bibitem{39}
O.~Garitselov, S.~P. Mohanty, E.~Kougianos, {Fast-Accurate Non-Polynomial Metamodeling for Nano-CMOS {PLL} Design Optimization}, in: V.~D. Agrawal, S.~T. Chakradhar (Eds.), 25th International Conference on {VLSI} Design, {IEEE} Computer Society, 2012, pp. 316--321.

\bibitem{40}
N.~Lourenço, E.~Afacan, R.~Martins, F.~Passos, A.~Canelas, R.~Póvoa, N.~Horta, G.~Dundar, {Using Polynomial Regression and Artificial Neural Networks for Reusable Analog IC Sizing}, in: 16th International Conference on Synthesis, Modeling, Analysis and Simulation Methods and Applications to Circuit Design, 2019, pp. 13--16.

\bibitem{41}
W.~Daems, G.~Gielen, W.~Sansen, {Simulation-based Generation of Posynomial Performance Models for the Sizing of Analog Integrated Circuits}, IEEE Transactions on Computer-Aided Design of Integrated Circuits and Systems 22~(5) (2003) 517--534.

\bibitem{44}
T.~Kiely, G.~Gielen, {Performance Modeling of Analog Integrated Circuits using Least-Squares Support Vector Machines}, in: Proceedings Design, Automation and Test in Europe Conference and Exhibition, Vol.~1, 2004, pp. 448--453 Vol.1.

\bibitem{42}
T.~S. Wu, C.~Alkan, T.~W. Chen, {Complexity Reduction for Analog Circuit Performance Models using Random Forests}, in: 17th IFIP International Conference on Very Large Scale Integration, IEEE, 2009, pp. 29--34.

\bibitem{43}
X.~Tang, A.~Xu, {Multi-Class Classification using Kernel Density Estimation on K-Nearest Neighbours}, Electronics Letters 52~(8) (2016) 600--602.

\bibitem{55}
A.~Lberni, M.~A. Marktani, A.~Ahaitouf, A.~Ahaitouf, {Analog Circuit Sizing based on Evolutionary Algorithms and Deep Learning}, Expert Systems with Applications 237 (2024) 121480.

\bibitem{46}
M.~Hassanpourghadi, S.~Su, R.~A. Rasul, J.~Liu, Q.~Zhang, M.~S.~W. Chen, {Circuit Connectivity Inspired Neural Network for Analog Mixed-Signal Functional Modeling}, in: 58th ACM/IEEE Design Automation Conference, 2021, pp. 505--510.

\bibitem{25}
Y.~Li, Y.~Wang, Y.~Li, R.~Zhou, Z.~Lin, {An Artificial Neural Network Assisted Optimization System for Analog Design Space Exploration}, IEEE Transactions on Computer-Aided Design of Integrated Circuits and Systems 39~(10) (2020) 2640--2653.

\bibitem{26}
S.~Du, H.~Liu, H.~Yin, F.~Yu, J.~Li, {A Local Surrogate-based Parallel Optimization for Analog Circuits}, AEU - International Journal of Electronics and Communications 134 (2021) 153667.

\bibitem{27}
A.~F. Budak, M.~Gandara, W.~Shi, D.~Z. Pan, N.~Sun, B.~Liu, {An Efficient Analog Circuit Sizing Method Based on Machine Learning Assisted Global Optimization}, IEEE Transactions on Computer-Aided Design of Integrated Circuits and Systems 41~(5) (2022) 1209--1221.

\bibitem{28}
S.~Yin, R.~Wang, J.~Zhang, X.~Liu, Y.~Wang, {Fast Surrogate-Assisted Constrained Multiobjective Optimization for Analog Circuit Sizing via Self-Adaptive Incremental Learning}, IEEE Transactions on Computer-Aided Design of Integrated Circuits and Systems 42~(7) (2023) 2080--2093.

\bibitem{29}
S.~Du, H.~Liu, Q.~Hong, C.~Wang, {A Surrogate-based Parallel Optimization of Analog Circuits using Multi-Acquisition Functions}, AEU - International Journal of Electronics and Communications 146 (2022) 154105.

\bibitem{30}
O.~Okobiah, S.~Mohanty, E.~Kougianos, {Fast Design Optimization Through Simple Kriging Metamodeling: A Sense Amplifier Case Study}, IEEE Transactions on Very Large Scale Integration Systems 22~(4) (2014) 932--937.

\bibitem{32}
X.~Li, L.~Chang, Y.~Cao, J.~Lu, X.~Lu, H.~Jiang, {Physics-Supervised Deep Learning–based Optimization with Accuracy and Efficiency}, Proceedings of the National Academy of Sciences 120~(35) (2023) e2309062120.

\bibitem{33}
J.~H. Holland, {Adaptation in Natural and Artificial Systems: An Introductory Analysis with Applications to Biology, Control, and Artificial Intelligence}, University of Michigan Press, Ann Arbor, 1975.

\bibitem{53}
V.~Bhatia, K.~Gupta, N.~Batra, N.~Pandey, Modelling a simple current to voltage converter using ann, in: IEEE 1st International Conference on Power Electronics, Intelligent Control and Energy Systems, 2016, pp. 1--4.

\bibitem{54}
E.~Dumesnil, F.~Nabki, M.~Boukadoum, {RF-LNA Circuit Synthesis using an Array of Artificial Neural Networks with Constrained Inputs}, in: IEEE International Symposium on Circuits and Systems, 2015, pp. 573--576.

\bibitem{35}
F.~Pedregosa, G.~Varoquaux, A.~Gramfort, V.~Michel, B.~Thirion, O.~Grisel, M.~Blondel, P.~Prettenhofer, R.~Weiss, V.~Dubourg, J.~Vanderplas, A.~Passos, D.~Cournapeau, M.~Brucher, M.~Perrot, E.~Duchesnay, {Scikit-learn: Machine Learning in {P}ython}, Journal of Machine Learning Research 12 (2011) 2825--2830.

\bibitem{37}
Y.~Yang, X.~Yin, D.~Chen, D.~Li, Y.~Yang, {The High-Efficiency Optimization Design Method for Two-Stage Miller Compensated Operational Amplifier}, IEEE Transactions on Circuits and Systems II: Express Briefs 71~(4) (2024) 2029--2033.

\end{thebibliography}

\end{document}